\documentclass{article}

\PassOptionsToPackage{numbers, sort&compress}{natbib}

\usepackage[preprint]{neurips_2026}

\usepackage{graphicx}
\usepackage{booktabs} 

\usepackage{hyperref}
\usepackage{xcolor}
\definecolor{royalblue}{rgb}{0.2549,0.4118,0.8824}

\usepackage{algorithm}
\usepackage{algpseudocode}

\hypersetup{
  colorlinks=true,
  linkcolor=royalblue,
  citecolor=royalblue,
  urlcolor=royalblue,
  filecolor=royalblue
}

\title{DBMol: Design of High-Affinity, Target-Specific Small Molecules through Structure Prediction Models}

\newcommand{\correspondence}{\thanks{Equal contribution. Correspondence to \texttt{yiming.qin@epfl.ch} and \texttt{kyi@mrc-lmb.cam.ac.uk}.}}

\author{%
  \makebox[0.30\textwidth][c]{Yiming Qin\correspondence} \\
  \makebox[0.30\textwidth][c]{EPFL, Switzerland}
  \And
  \makebox[0.30\textwidth][c]{Kai Yi\footnotemark[1]} \\
  \makebox[0.30\textwidth][c]{MRC-LMB, UK}
  \And
  \makebox[0.30\textwidth][c]{Miruna Cretu} \\
  \makebox[0.30\textwidth][c]{University of Cambridge, UK}
  \AND
  \makebox[0.30\textwidth][c]{Sjors H.W. Scheres} \\
  \makebox[0.30\textwidth][c]{MRC-LMB, UK}
  \And
  \makebox[0.30\textwidth][c]{Pietro Liò} \\
  \makebox[0.30\textwidth][c]{University of Cambridge, UK}
  \And
  \makebox[0.30\textwidth][c]{Pascal Frossard} \\
  \makebox[0.30\textwidth][c]{EPFL, Switzerland}
}

\usepackage{amsmath,amsfonts,bm}

\def\eqref#1{equation~\ref{#1}}

\def\1{\bm{1}}

\DeclareMathAlphabet{\mathsfit}{\encodingdefault}{\sfdefault}{m}{sl}
\SetMathAlphabet{\mathsfit}{bold}{\encodingdefault}{\sfdefault}{bx}{n}

\usepackage{lipsum}

\usepackage{siunitx} 
\usepackage{color, colortbl}
\definecolor{Gray}{gray}{0.9}
\definecolor{LightGray}{gray}{0.97}
\newcolumntype{g}{>{\columncolor{LightGray}}S}

\usepackage{bm}

\usepackage{xcolor}  
\usepackage{multicol}
\usepackage{enumitem}

\usepackage{minitoc}
\doparttoc 

\usepackage{amsfonts}

\usepackage{setspace}

\usepackage{float}
\usepackage{algorithm}
\usepackage{subcaption}  
\usepackage{nicefrac}
\usepackage{wrapfig}

\usepackage{etoolbox}

\usepackage{amssymb}
\usepackage{amsthm}
\usepackage{amsmath}
\usepackage{graphicx}
\theoremstyle{plain}

\theoremstyle{definition}

\theoremstyle{remark}

\newcommand{\ca}[1]{\textcolor{gray}{\small #1}}

\usepackage[capitalize,noabbrev]{cleveref}
\Crefname{equation}{Eq.}{Eqs.} 
\Crefname{section}{Sec.}{Secs.}
\Crefname{table}{Tab.}{Tabs.}
\Crefname{algorithm}{Alg.}{Algs.}

\begin{document}

\maketitle

\newcommand{\kai}[1]{\textcolor{purple}{#1}}
\doparttoc 
\faketableofcontents 

\begin{abstract}
Designing small molecule ligands that bind with high affinity to specific protein pockets is a fundamental goal in drug discovery, as small molecules constitute a major fraction of approved therapeutics.
Recent breakthroughs in structure prediction, such as AlphaFold-3 and Boltz-2, enable accurate biomolecular interaction prediction and show promise as foundation models for downstream tasks, including binding affinity prediction. We propose to leverage these models and introduce DBMol, a new structure predictor-guided framework for de novo small molecule design. 
DBMol formulates an alternating optimization and projection process. In the optimization stage, DBMol starts from an initial molecule and uses gradient-based optimization to improve pocket-specific interactions and predicted binding affinity using a structure prediction model. In the projection stage, a flow-matching model maps the optimized molecular graph to discrete and chemically valid molecules.
Experiments show that DBMol effectively optimizes the Boltz-2 affinity proxy and generates molecules with strong predicted affinity and specificity under Boltz-2 evaluation.
To reduce self-confirmation bias, we further evaluate generated molecules using held-out metrics, including AF3-based evaluation.
DBMol substantially improves pocket coverage while maintaining molecular diversity over unconditional generation, and is competitive under held-out metrics despite the absence of reference-ligand supervision.
These results support the promise of structure prediction models as effective optimization signals for de novo molecular design.
\end{abstract}

\section{Introduction}
\label{sec:intro}
Small-molecule modulators that target specific protein pockets with high affinity are central to modern therapeutics. G protein-coupled receptors (GPCRs), for instance, represent one of the most important drug target families. More than 30\% of FDA-approved drugs act on GPCRs~\cite{hauser2018pharmacogenomics}, with the objective of either activating or inhibiting receptor signaling by binding to specific conformational pockets. Achieving both high binding affinity and strong target specificity is therefore a core objective in molecule design, as it directly determines therapeutic efficacy and off-target risk.
\begin{figure*}[t]
    \centering
    \begin{subfigure}[b]{0.99\linewidth}
            \includegraphics[width=\linewidth]{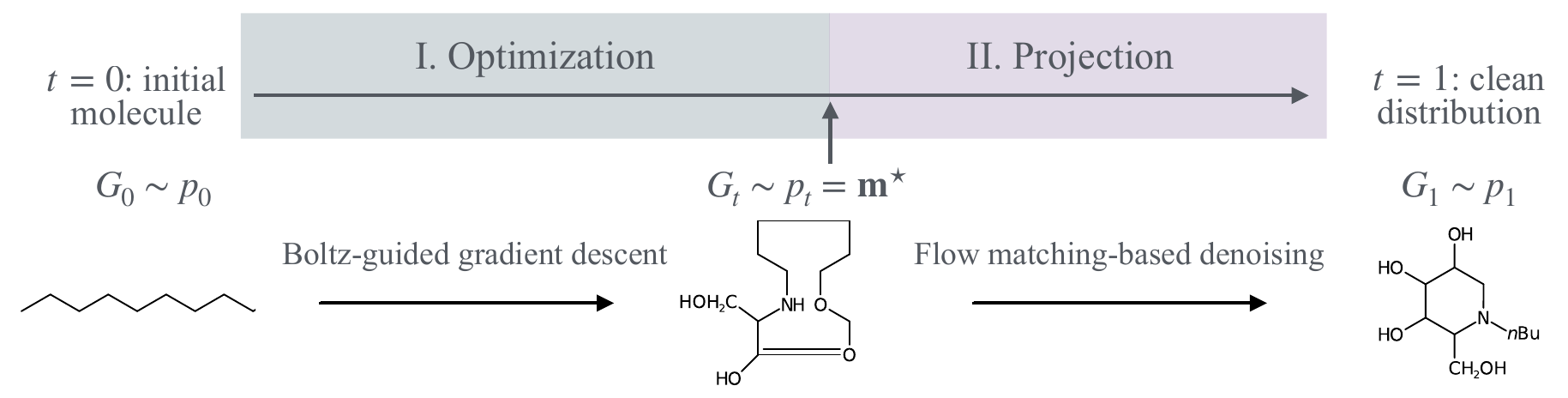}
    \end{subfigure}
    \caption{
    \textbf{Framework of DBMol}. (I) \textbf{Optimization}, where Boltz-2-conditioned gradient updates start from $G_0$, an uninformative initial molecule graph visualized here as a simple linear carbon chain, and refine relaxed molecule representation to improve predicted binding affinity and pocket specificity for a given protein pocket; and (II). \textbf{Projection}, where a flow-matching-based molecule denoising model maps a noisy molecule $G_t$ (in its optimized representation $\mathbf{m}^\star$) onto discrete, chemically valid molecule graphs $G_1$.
    } 
    \label{fig:main}
    \vspace{-15pt}
\end{figure*}
Existing ligand design methods can be broadly divided into two categories. Structure-based, pocket-aware generators, such as DiffSBDD~\cite{schneuing2024structurebaseddrugdesignequivariant}, FLOWR~\cite{Cremer2025FLOWRFM}, and Pocket2Mol~\cite{Peng2022Pocket2MolEM}, explicitly condition ligand generation on protein pocket geometry.
In contrast, ligand-centric models \citep{molsnapper, adamsShEPhERD2024, gao2024generative, shen2025compositional, rekesh2025syncogen} guide generation using docking objectives, reference ligands, or pharmacophore profiles.
These two directions therefore face complementary limitations: pocket-aware generators are constrained by the coverage of currently available protein--ligand complexes, while ligand-centric optimization methods depend on target-specific objectives, known ligands, or costly retraining procedures, restricting their generalization to new proteins and binding sites.
This motivates an alternative way of leveraging protein information: rather than learning ligand generation only from curated complexes or optimizing task-specific ligand scores, we can use modern structure prediction models as general-purpose protein--ligand interaction signals.

Recent advances in biomolecular structure prediction make this increasingly promising.
AlphaFold-3~\cite{abramson2024accurate} shows accurate modeling of multi-molecule complexes, including protein--ligand interactions, enabling structure-based in-silico ligand design workflows that were previously impractical at scale. Building on this progress, Boltz-2~\cite{passaro2025boltz} extends such models to both continuous binding strength and binder probability from protein--ligand inputs.
Motivated by these developments, we propose DBMol, a small-molecule design framework that uses structure prediction models to generate ligands specific to a target protein pocket.
Unlike conventional generative models, DBMol does not require curated protein--ligand complex datasets, known ligands for the target, or task-specific retraining.
Instead, it directly leverages differentiable signals from a structure prediction model, such as Boltz-2~\cite{passaro2025boltz}, to guide molecule design through an alternating optimization and projection procedure, as illustrated in \cref{fig:main}.
As a result, DBMol enables flexible de novo small-molecule generation conditioned only on a target protein sequence and target pocket positions.

More specifically, DBMol first performs an optimization step (stage I in \cref{fig:main}), where a differentiable interaction objective derived from the structure prediction model is optimized to improve predicted binding affinity, pocket specificity, and predicted complex confidence.
This objective provides gradient signals that guide molecule representations toward structures aligned with the target protein pocket.
However, directly applying differentiable guidance to small molecules is challenging because atom types, formal charges, and bonding patterns are tightly coupled and must satisfy strict chemical constraints.
Naïve gradient-based updates can therefore distort the molecular distribution and frequently produce invalid or chemically implausible structures.
This necessitates an explicit projection mechanism onto the discrete, chemically valid molecule manifold.
DBMol therefore uses a diffusion-based molecule completion model, such as DeFoG~\cite{qin2025defog}, as the projection stage (stage II in \cref{fig:main}).
This stage maps optimized but potentially invalid molecule representations back to chemically valid 2D molecule graphs through iterative denoising, while preserving the optimized structural signal.
Synthesizability is then encouraged by mapping each generated molecule to a structurally similar, synthetically accessible compound~\cite{gao2024generative}.

We evaluate DBMol on LIT-PCBA~\cite{litpcba}, a benchmark for target-specific molecular generation.
DBMol effectively optimizes the Boltz-2 proxy and substantially improves over unconditional generation under structure-predictor-aligned metrics.
These results show that the proposed optimization-and-projection pipeline can capture useful structure-predictor signals for pocket-specific molecule design.
Since DBMol is optimized using Boltz-based signals, we further evaluate generated molecules using held-out AF3-based metrics as a complementary structural assessment, rather than treating the optimization proxy as ground truth.
Under this held-out evaluation, DBMol achieves substantially higher target-pocket coverage while preserving molecular diversity, and remains competitive with reinforcement-learning-based and conditional generative baselines despite using weaker supervision: no reference ligands, curated pocket--ligand training pairs, or task-specific retraining.
Beyond the benchmark setting, we further show that DBMol can be applied to new protein pockets without known ligands, highlighting its intended role as a flexible de novo design framework that can benefit from continued advances in biomolecular structure prediction models.

\vspace{-5pt}
\section{Related Work}
\label{sec:related_work}
\vspace{-5pt}

\paragraph{Generative models for molecule design.}
Molecular generative models can be broadly categorized into three classes: SMILES-based sequence models, graph-based 2D models, and 3D structure-aware models.
SMILES-based models treat molecules as sequences using autoregressive or transformer-based architectures~\cite{GmezBombarelli2016AutomaticCD, bagal2022molgpt}.
Graph-based 2D models represent molecules as atom–bond graphs and have been extensively studied, using variational autoencoders~\cite{pmlr-v70-kusner17a, liu2018constrained, jin2019junctiontreevariationalautoencoder}, normalizing flows~\cite{zang2020moflow}, and diffusion-based methods~\cite{vignac2022digress}. Discrete flow-matching frameworks such as DeFoG~\cite{qin2025defog} further improve generation efficiency while preserving strong performance.
3D structure-aware models aim to jointly generate molecule topology and three-dimensional atomic coordinates, typically using E(3)-equivariant architectures with diffusion or flow matching~\cite{hoogeboom2022equivariant,vignac2023midi,irwin2025semlaflowefficient3d}, or keeping a non-equivariant architecture~\cite{joshi2025allatom, vonessen2025tabascofastsimplifiedmodel}.
Recently, there has been increased attention to generating molecules that, beyond accurate 2D topology and geometric fidelity, are constrained to synthesizable space~\cite{cretu2025synflownet,koziarski2024rgfn, shen2025compositional, gao2024generative, rekesh2025syncogen}.
In this work, we focus on 2D molecule graph generation, which provides a discrete representation of molecule structure while aligning naturally with the Boltz-2 input format, and we incorporate a synthesis-aware pipeline for practical molecule design.

\paragraph{Structure prediction models.}
Structure prediction models have moved beyond single-protein folding and are increasingly used as foundation models for biomolecular systems. AlphaFold-3~\cite{abramson2024accurate} showed that diffusion-based all-atom models can predict multi-component complexes, including protein-ligand interactions and nucleic-acid assemblies. Recent open-source efforts further extend structure prediction toward functional readouts such as binding affinity and conformational change. For example, Boltz-2~\cite{passaro2025boltz} reports protein-ligand affinity predictions competitive with physics-based free-energy methods such as FEP, while being substantially faster, making these models attractive for virtual screening and reducing reliance on expensive simulations.
Structure prediction models have also been explored for inverse design, primarily in the context of protein sequence and structure optimization. Representative approaches include all-atom sequence design with explicit ligands and ions~\cite{yi2025allatom}, diffusion-time hallucination for de novo protein discovery~\cite{cho2025protein}, and gradient-based optimization through differentiable structure predictors for protein–protein or protein–nucleic acid interfaces~\cite{pacesa2025one,team2025pxdesign,cho2025boltzdesign1}.
However, despite these advances, a comparable differentiable, structure-predictor-guided framework for \emph{small-molecule} design remains largely unexplored.

\vspace{-5pt}

\section{DBMol}\label{sec:method_graphs}
\vspace{-5pt}

In this section, we present DBMol, an optimization–projection framework for structure-guided molecule design. We first define the problem setting (\cref{sec:notations}), then describe how gradients from the structure predictor Boltz-2 are used to optimize a relaxed molecular representation (\cref{sec:boltz}). Next, we introduce a discrete flow–matching projection step to recover chemically valid molecular graphs (\cref{sec:denoising}). Finally, we present a refinement stage that enforces synthesizability (\cref{sec:synformer}).

\vspace{-5pt}
\subsection{Problem Setting and Molecular Representation}
\label{sec:notations}

We study \emph{structure-predictor-aligned molecule design}, which can be viewed as a conditional generation problem guided by a differentiable surrogate objective. Given a protein sequence and a predefined binding pocket, we treat a structure prediction model (e.g., Boltz-2~\cite{passaro2025boltz}) as a differentiable scoring function and optimize molecules to improve predicted protein-ligand interactions.
Importantly, we do not assume access to ground-truth binding affinities, and we do not treat the structure predictor as a physical oracle; instead, it serves as a proxy objective whose inductive biases guide molecule design.

\paragraph{Molecular representation.}

We represent a molecule as an undirected graph with $N$ nodes,
\begin{equation}
G = \bigl( x_{1:N}, c_{1:N}, e_{1:i<j:N} \bigr),
\end{equation}
where $x_{1:N} = (x^{(n)})_{1 \le n \le N}$ denote atom types,
$c_{1:N} = (c^{(n)})_{1 \le n \le N}$ denote formal charges,
and $e_{1:i<j:N} = (e^{(ij)})_{1 \le i < j \le N}$ denote bond types.
Each atom type satisfies $x^{(n)} \in \mathcal{X}$, each charge $c^{(n)} \in \mathcal{C}$, and each bond type $e^{(ij)} \in \mathcal{E}$, where one category in $\mathcal{E}$ explicitly represents the absence of a bond.
Following prior work~\cite{vignac2022digress} on molecule graph generation, we assume a fully connected graph, allowing bonded and non-bonded atom pairs to be modeled within a unified categorical space. We adopt this discrete notation for the projection stage in \cref{sec:denoising}.

\begin{wrapfigure}{l}{0.57\textwidth}
\vspace{-20pt}
\centering
\begin{minipage}{0.57\textwidth}
\begin{algorithm}[H]
\caption{Boltz-guided optimization (\cref{sec:boltz})}
\label{alg:dbmol-em}
\begin{algorithmic}[1]
\State \textbf{Input:} pocket $\mathcal{P}$, Boltz2, DeFoG denoiser $f_{\theta}$
\State \textbf{Hyperparams:} weights $(w_1,w_2)$, optimization steps $K$, late start $t$, step size $\eta$
\State \textbf{Init:} simple discrete molecule $G_0$
\State $\mathbf{m}_0 \gets \operatorname{Relax}(G_0)$ \Comment{\ca{atom/bond probs + charges}}
\For{$k = 0$ \textbf{to} $K-1$}
    \State $\mathcal{L} \gets 
    \mathcal{L}_{\text{aff}}(\mathbf{m}_k) +
    w_1\mathcal{L}_{\text{contact}}(\mathbf{m}_k) +
    w_2\mathcal{L}_{\text{conf}}(\mathbf{m}_k)$
    \State $\mathbf{g}_k \gets \nabla_{\mathbf{m}_k} \mathcal{L}(\mathbf{m}_k)$
    \State $\mathbf{m}_{k+1} \gets \operatorname{Update}(\mathbf{m}_{k},\,\mathbf{g}_k,\,\eta)$
\EndFor
\State $\mathbf{m}^\star  \gets  \mathbf{m}_K$
\State $G_t \sim p(\cdot \mid \mathbf{m}^\star)$
\State $G_1 \gets \operatorname{Denoise}(G_t, f_{\theta}, t)$
\State \Return $G_1$
\end{algorithmic}
\end{algorithm}
\end{minipage}
\vspace{-10pt}
\end{wrapfigure}
To enable gradient-based optimization in our framework, we adopt a \emph{continuous relaxation} of the discrete molecule graph.
Specifically, atom types $x_{1:N}$ and bond types $e_{1:i<j:N}$ are represented as categorical probability vectors on the probability simplex, while formal charges $c_{1:N}$ are represented as continuous values to align with Boltz-2.
We denote this relaxed continuous representation by $\mathbf{m}$, and distinguish it from the discrete molecule graph $G$.
This relaxation does not carry a strict probabilistic interpretation over molecules, nor does it define a generative distribution; rather, it serves as a parameterization that enables constrained gradient-based design using differentiable structure prediction models. We adopt this relaxed notation $\mathbf{m}$ for \cref{sec:boltz} (summarized in Alg. \ref{alg:dbmol-em}).

We next describe the structure-predictor-guided optimization of DBMol under the notations above.
\vspace{-5pt}

\subsection{Optimization through Structure Prediction Models}
\label{sec:boltz}

We leverage a structure prediction model to guide molecule design through gradient-based optimization.
In particular, we adopt Boltz-2~\cite{passaro2025boltz} as a structure predictor that provides differentiable signals characterizing protein-ligand interactions, including binding affinity estimates and structural confidence measures.
These signals define a fixed objective for optimization; DBMol does not perform model retraining, reinforcement learning, or inference over learned molecule distributions.

\paragraph{Objective.}
Given a relaxed molecule representation $\mathbf{m}$, we define a composite objective
\begin{equation}
\mathcal{L}(\mathbf{m})=
\mathcal{L}_{\mathrm{affinity}}(\mathbf{m})
+
w_1 \mathcal{L}_{\mathrm{contact}}(\mathbf{m})
+
w_2 \mathcal{L}_{\mathrm{conf}}(\mathbf{m}),
\end{equation}
where $\mathcal{L}_{\mathrm{affinity}}$ combines the minimization of Boltz-2 affinity regression value with maximization of the binder/non-binder classification logit from the structure prediction model, biasing generation toward stronger binders.
$\mathcal{L}_{\mathrm{contact}}$ encourages interactions with residues in the predefined binding pocket, while $\mathcal{L}_{\mathrm{conf}}$ aggregates Boltz-2 confidence signals to encourage plausible complex geometries.
All terms are written as losses, so lower values are preferred: higher-is-better confidence scores are negated, and error-like quantities are used as penalties. The positive weights $(w_1, w_2)$ balance these objectives, with details provided in App. \ref{app:boltz_setup}. 
Overall, the objective encourages strong, target-specific binders with high-confidence predicted interfaces.
While continuous relaxation enables efficient gradient-based optimization, directly discretizing the optimized representation $\mathbf{m}$ almost always produces invalid or chemically implausible molecular graphs. This occurs because atom types and bond configurations are tightly coupled, and these constraints are not enforced in the relaxed space.
To address this issue, we introduce a \emph{projection step} that maps noisy or invalid proposals to discrete molecular graphs that satisfy basic chemical validity constraints.
\begin{figure*}[t]
    \centering
    \begin{subfigure}[b]{1.0\linewidth}
        \centering
        \begin{subfigure}[b]{1.0\linewidth}
        \includegraphics[width=\linewidth]{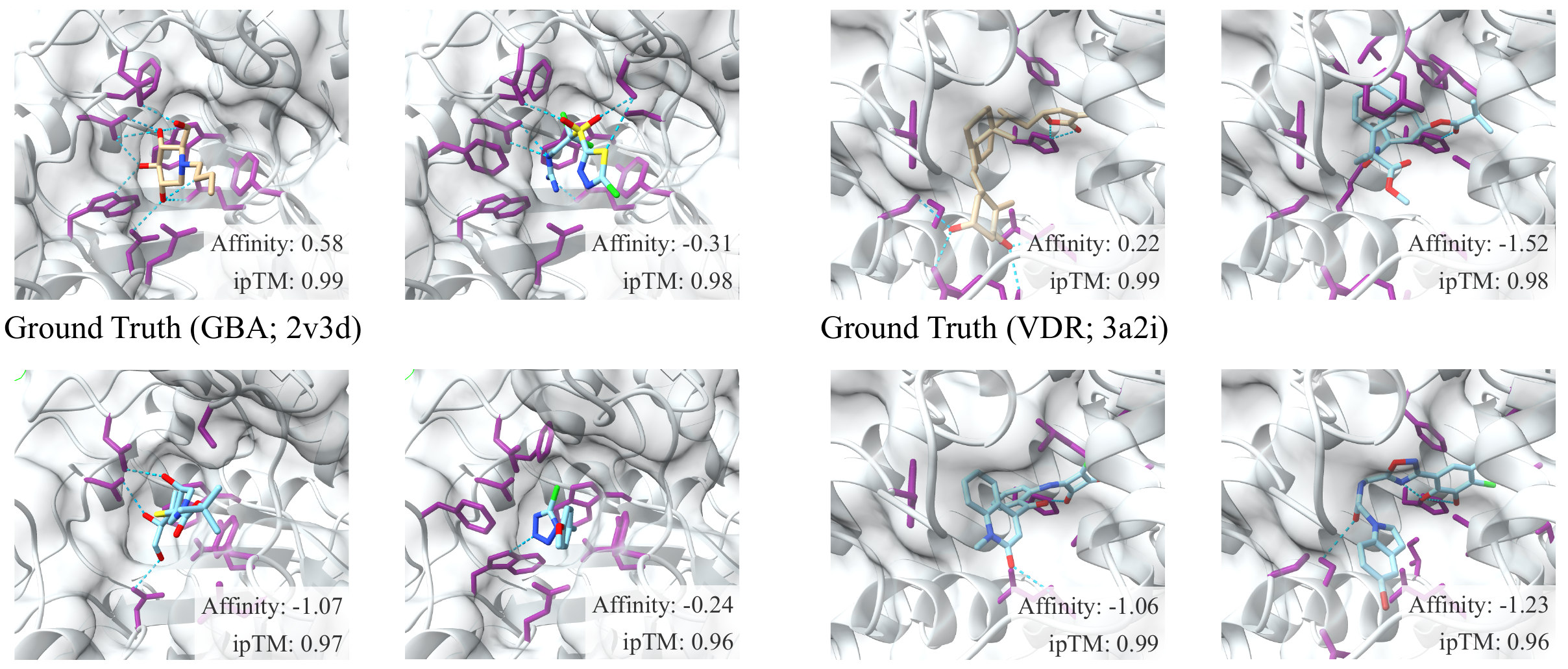}
        \end{subfigure}
    \end{subfigure}
    \caption{\textbf{Protein–ligand interactions predicted by Boltz-2.} For each target, the reference ligand and its interaction pattern are shown in beige in the upper left, while three molecules found via DBMol after SynFormer post-processing are shown in blue. The protein backbone and molecular surface are shown in gray, interacting protein side chains are highlighted in purple, and hydrogen bonds are shown as dashed light-blue lines. Across both proteins, the generated molecules bind within the target pocket and exhibit coherent predicted binding poses. The predicted binding affinities and ipTM scores are reported in the bottom right of each subfigure.
    }
    \vspace{-10pt}
    \label{fig:cifs}
\end{figure*}

\paragraph{Contact loss.}
To encourage pocket-specific binding, we add a contact loss based on the distogram predicted by the structure prediction model. For each pocket residue $i$ and ligand token $j$, the model outputs logits over $D$ distance bins. After softmax, we obtain $p_{ij}(d)$ and define
\begin{equation}
\mathcal{L}_{\mathrm{pair}}(i,j)
= -\log\!\left(\sum_{d<d_{\mathrm{cutoff}}} p_{ij}(d) + \epsilon\right),
\end{equation}
where $\epsilon$ is a small constant and $d_{\mathrm{cutoff}}=8.0$ \AA. We compute this loss only over pocket residues $\mathcal{P}$. For each residue $i$, we select the ligand atom with the strongest predicted contact, $j_i^\star=\arg\min_j \mathcal{L}_{\mathrm{pair}}(i,j)$, and define
\begin{equation}
\mathcal{L}_{\mathrm{pocket}}(\mathbf{m})
=
\frac{1}{|\mathcal{P}|}
\sum_{i\in \mathcal{P}}
\mathcal{L}_{\mathrm{pair}}(i,j_i^\star).
\end{equation}
Simultaneously enforcing many pocket--ligand contacts can make optimization unstable and lead to noisy gradients. We therefore add an anchoring term based on the single strongest pocket residue--ligand atom pair:
\begin{equation}
(i^\star,j^\star)=\arg\min_{i\in\mathcal{P},j}\mathcal{L}_{\mathrm{pair}}(i,j),
\mathcal{L}_{\mathrm{anchor}}(\mathbf{m})
=
\mathcal{L}_{\mathrm{pair}}(i^\star,j^\star).
\end{equation}
The final contact loss averages the pocket-level and anchor-level terms:
\begin{equation}
\mathcal{L}_{\mathrm{contact}}(\mathbf{m})
=
\frac{1}{2}\mathcal{L}_{\mathrm{pocket}}(\mathbf{m})
+
\frac{1}{2}\mathcal{L}_{\mathrm{anchor}}(\mathbf{m}).
\end{equation}
The anchor term provides a stable global pull toward the target pocket, while the pocket-level term encourages broader local contacts, improving stability and reducing diffuse interaction patterns.

\paragraph{Gradient updates in the relaxed space.}
Direct optimization over discrete molecule graphs is infeasible due to combinatorial constraints.
We therefore perform gradient-based optimization in the relaxed molecule space defined in the previous subsection.
At iteration $k$, we compute gradients of the objective with respect to the relaxed representation of the molecule $\mathbf{m}_k$,
\begin{equation}
\mathbf{g}_k = \nabla_{\mathbf{m}_k} \mathcal{L}(\mathbf{m}_k),
\end{equation}
and apply constrained gradient descent to preserve valid categorical simplices.
For atom and bond type variables, under step size $\eta$, the update takes the form
\begin{align}
\tilde{\mathbf{m}}_{k+1} = \mathrm{clip}(\mathbf{m}_k - \eta \mathbf{g}_k, 0, 1)\rightarrow
\mathbf{m}_{k+1} = \frac{\tilde{\mathbf{m}}_{k+1}}{\sum_l \tilde{\mathbf{m}}_{k+1}^{(l)}}.
\end{align}
Here, clipping prevents negative or overly large category weights, while normalization ensures that each atom-type or bond-type vector sums to one.
For continuous charge variables, we apply additive gradient-descent updates followed by clipping to a predefined range.
A detailed description of how gradients affect atom types, bond types, and atomic charges is provided in App. \ref{app:boltz_setup}.

While these operations implement constrained gradient descent in the relaxed space, they do not ensure chemical validity. We therefore introduce a separate projection (denoising) step in \cref{sec:denoising}.

\subsection{Projection to Chemically Valid Molecules}
\label{sec:denoising}

\paragraph{Learned projection via discrete denoising.}
We instantiate this projection using a denoising model based on discrete flow matching, specifically DeFoG~\cite{qin2025defog}.
Discrete flow matching formulates denoising as a continuous-time process over discrete states, enabling gradual and controlled corrections of atom, bond, and charge types.
This property makes it particularly well suited for projection, as local chemical inconsistencies can be resolved without overwriting the global structural signal introduced during optimization.
Full technical details of the discrete flow-matching formulation and training procedure are provided in App.~\ref{app:dfm}.

DeFoG is trained only on chemically valid molecular graphs and learns to recover chemically consistent structures from corrupted graph inputs.
We do not interpret this denoising process as probabilistic inference; instead, we use it as a learned projection operator that enforces chemical validity while preserving the structural signal introduced by optimization stage in \cref{sec:boltz}.

Given an optimized relaxed representation $\mathbf{m}^\star$, we first instantiate a noisy discrete molecule graph $G_t$, by sampling the atom and bond types from $\mathbf{m}^\star$ and rounding the continuous charges to their nearest integers. Since charges are encoded as scalars in $\mathbf{m}^\star$ in Boltz-2, we discretize them in $G_t$ by rounding to the nearest integer.
This sampled graph may violate chemical constraints, and we therefore apply the denoising model to project $G_t$ back onto a valid molecule graph $G_1$.

\vspace{-5pt}
\paragraph{Late-time denoising.}
A key design choice is the initialization point of the denoising process in the flow-matching trajectory.
Starting from an early denoising time introduces too much randomness and can wash out the optimization signal from the structure predictor. Starting from an overly late time, however, produces many invalid structures that the denoiser cannot reliably repair. We therefore initialize denoising at a \emph{late diffusion time} $t$, which allows the model to ensure sufficient chemical validity while largely preserving the optimized molecule structure.
In practice, we set $t = 0.8$ in all experiments, which provides a favorable trade-off.

\begin{table*}[t]
    \centering
    \footnotesize
    \caption{\textbf{Mean molecular generation results averaged across the seven LIT-PCBA targets.}
    Rank is computed over all held-out metrics except Boltz2 Success using standard competition ranking. DBMol results are reported from 100 valid molecules without any post-selection.
    Gray rows correspond to our method, and the best average rank and Boltz2 Success are highlighted in bold.}
    \resizebox{0.99\linewidth}{!}{
    \begin{tabular}{@{}llcccccc>{\columncolor{Gray}}c>{\columncolor{Gray}}c@{}}
    \toprule
    Model & Condition 
    & BCov $\uparrow$ 
    & AF3 Succ. $\uparrow$ 
    & iPAE $\downarrow$ 
    & Dist $\downarrow$ 
    & Div. $\uparrow$ 
    & Vina $\downarrow$ 
    & Rank $\downarrow$
    & \ \ \ Boltz Succ. $\uparrow$ \\
    \midrule
    DeFoG & Unconditional 
    & 0.39 & 0.89 & 11.7 & 4.6 & 0.89 & -8.3 & 5.50 & 0.13 \\
    \midrule
    SynFormer & Ligand 
    & 0.55 & 0.93 & 10.8 & 3.8 & 0.87 & -8.9 & \textbf{1.83} & 0.47 \\
    ShEPhERD & Pharmacophore
    & 0.42 & 0.91 & 10.8 & 4.2 & 0.87 & -8.5 & 4.00 & 0.20 \\
    SynCoGen & Pharmacophore
    & 0.51 & 0.92 & 11.6 & 4.5 & 0.50 & -10.3 & 3.67 & \textbf{0.62} \\
    CGFlow-ZS & 3D pocket
    & 0.46 & 0.92 & 10.9 & 4.5 & 0.78 & -10.4 & 3.33 & 0.32 \\
    DiffSBDD & 3D pocket
    & 0.50 & 0.94 & 10.9 & 4.6 & 0.78 & -8.8 & 3.67 & 0.44 \\
    \rowcolor{Gray}
    DBMol & Pocket
    & 0.51 & 0.92 & 11.8 & 4.6 & 0.89 & -8.7 & 3.83 & 0.47 \\ 
    \bottomrule
    \end{tabular}}
    \label{tab:main_results}
    \vspace{-10pt}
\end{table*}

\vspace{-5pt}
\subsection{Synthesizability-Aware Design}
\label{sec:synformer}
Chemical validity does not necessarily imply practical synthesizability: a generated molecule may satisfy valence constraints while still being unstable, implausible, or difficult to synthesize. 
We therefore include an optional synthesis-aware post-processing step based on SynFormer~\cite{gao2024generative}. 
Given a molecule generated by DBMol, SynFormer maps it to a structurally similar molecule that is reachable through chemically valid reaction pathways. 
This step is used only to assess whether the target-specific signal discovered by DBMol can be preserved after moving molecules toward synthesizable chemical space; it is not part of the Boltz-guided optimization objective. 
In our synthesis-aware variant, this mapping is applied independently to each DBMol molecule, without Boltz-2-, AF3-, Vina-, or docking-based filtering.

\vspace{-5pt}

\section{Experiments}
\label{sec:exp}
\vspace{-5pt}

We evaluate whether DBMol can use structure prediction models as optimization signals for \textit{de novo} molecule design.
Our experiments have three goals:
(i) to test whether DBMol optimizes the Boltz-2 proxy, as assessed via both Boltz-2- and AF3-based evaluation on seven targets from the LIT-PCBA benchmark;
(ii) to investigate whether this framework extends to \textit{de novo} targets beyond the original benchmark;
and (iii) to validate the main design choices through ablations.

\subsection{Experimental Setup}

\paragraph{Baselines.}
We compare DBMol with representative baselines across conditioning and supervision regimes.
DeFoG is an unconditional graph-generation reference.
SynFormer~\cite{gao2024generative} uses reference-ligand guidance; ShEPhERD~\cite{adamsShEPhERD2024} and SynCoGen~\cite{rekesh2025syncogen} use pharmacophore guidance; CGFlow~\cite{shen2025compositional} and DiffSBDD~\cite{schneuing2024structurebaseddrugdesignequivariant} are pocket-conditioned structure-based generators.
DiffSBDD conditions on 3D pocket information, typically from a broader $8$~\AA~\  pocket region, whereas DBMol only uses pocket residue positions within $3.5$~\AA~\, 
and does not condition on a fixed pocket conformation, and therefore relies on less structural but more flexible information.
Together, these baselines cover the main regimes relevant to target-specific molecule generation, including no target information, reference ligands, pharmacophores, learned pocket rewards, and pocket--ligand training data.
This comparison positions DBMol as a structure-predictor-guided approach that remains competitive under weaker supervision, suggesting a promising path toward flexible de novo molecular design.

\paragraph{Evaluation protocol.}
For the main benchmark, we report 100 molecules per target for each method.
For the projection stage, we generate 2048 candidates, discard chemically invalid projected graphs, and randomly sample 100 valid molecules, which takes around 10 seconds in total.
This step enforces only chemical validity and does not use any evaluation metrics for selection.

\paragraph{Metrics.}
We evaluate all methods using held-out structural metrics, proxy-aligned reference metrics, and chemical-quality metrics, with details in App.~\ref{app:eval}.
Our primary evaluation uses \emph{AlphaFold3-based held-out metrics}, since AlphaFold3 is not used during DBMol optimization.
These include BindCov (denoted as BCov as an abbreviation), measuring AF3-predicted pocket coverage; AF3 Succ., a combined success rate based on sufficient AF3 ipTM and successful pocket binding; iPAE, the AF3-predicted interface error; and Dist, the mean minimum ligand-to-pocket distance.
We also report AutoDock Vina scores and molecular diversity, measured by average pairwise ECFP4 Tanimoto distance, both used only for evaluation.
We additionally report \emph{Boltz2 Success}, defined by favorable Boltz-2 affinity and sufficient pocket specificity.
Since Boltz-2 provides the optimization signal, Boltz2 Success is treated only as a diagnostic proxy-aligned reference metric rather than the primary basis for comparison.

\subsection{Pocket-Specific Molecule Design on LIT-PCBA}

We evaluate DBMol on the seven-target LIT-PCBA protocol used in work~\cite{rekesh2025syncogen}.
For DBMol, we use one fixed configuration across all LIT-PCBA targets, including loss weights, optimization hyperparameters, and 50 optimization iterations.
This avoids target-specific iteration selection or hyperparameter tuning.
The full setup is reported in App.~\ref{app:experimental_details}. We use the ground-truth ligand size only to fix the number of atoms; no reference-ligand geometry, or interaction pattern is used.

\textit{We first ask whether the optimization-projection framework improves the proxy.}
The comparison in \cref{tab:main_results} shows that DBMol effectively exploits structure-predictor guidance.
Under the proxy-aligned Boltz2 Success metric, DBMol improves over unconditional generation from $0.13$ to $0.47$, indicating that projection preserves useful optimization signals after mapping relaxed representations back to discrete molecule graphs.

\textit{We next ask whether Boltz-guided optimization transfers to held-out evaluation.}
DBMol achieves strong held-out performance despite using weaker supervision than several baselines.
Using only the target pocket and the structure predictor, DBMol obtains a BCov of $0.51$, an AF3 Success of $0.92$, and high diversity of $0.89$.
Its average rank over held-out metrics is $3.83$, which is competitive with pocket-conditioned and pharmacophore-guided methods.
SynFormer achieves the best average rank and the strongest BCov and Dist, but it relies on the strongest conditioning setting, where a ground-truth ligand is available; CGFlow-ZS achieves a very low Vina score, supported by its training reward encouraging a proxy of this metric.
Nevertheless, DBMol remains competitive under weaker supervision and achieves the highest diversity among target-aware methods.
These results suggest that the Boltz-guided optimization signal transfers to independent AF3-based evaluation metrics, supporting DBMol as a flexible structure-predictor-guided design framework.

\paragraph{Proxy tradeoff analysis.}
We further analyze the proxy tradeoff by comparing DBMol with a less confidence-focused variant that optimizes only the ipTM confidence term for $L_\text{conf}$.
This variant increases Boltz2 Success from $0.47$ to $0.69$, surpassing the strongest baseline score of $0.62$.
Both this variant and DBMol reach a high $ipTM$ of $0.85$, indicating similar confidence under this metric.
This confirms that DBMol can strongly optimize the Boltz-2 proxy, especially affinity and pocket specificity.
However, this proxy gain comes with worse held-out metrics: BCov drops from $0.51$ to $0.46$, iPAE increases from $11.8$ to $12.3$, Dist increases from $4.6$ to $4.7$, and Vina weakens from $-8.7$ to $-8.3$.
AF3 Success also slightly decreases from $0.92$ to $0.91$.
These results show that stronger proxy optimization alone is insufficient, and that a more balanced confidence objective is important for improving held-out transfer.

\paragraph{Synthesizability.}
\begin{wraptable}[11]{r}{0.75\textwidth}
    \centering
    \vspace{-10pt}
    \footnotesize
    \caption{\textbf{Effect of SynFormer post-processing.}
    We compare DBMol candidates before and after SynFormer-based synthesizability post-processing. \textit{Retained} reports the average number of molecules kept per target after post-processing or size filtering. We abbreviate \textit{size filter} with \textit{sf}.}
    \setlength{\tabcolsep}{2.4pt}
    \renewcommand{\arraystretch}{1.05}
    \resizebox{1.0\linewidth}{!}{
    \begin{tabular}{@{}lccccccc>{\columncolor{Gray}}c@{}c}
    \toprule
    Setting
    & Retained
    & BCov $\uparrow$
    & AF3 Succ. $\uparrow$
    & iPAE $\downarrow$
    & Dist $\downarrow$
    & Vina $\downarrow$
    & Div $\uparrow$
    & Avg. Rank $\downarrow$ 
    & \ \ Syn. $\uparrow$\\
    \midrule
    DBMol
    & 100 & 0.51 & 0.92 & 11.8 & 4.6 &-8.7 & 0.89 &  \textbf{1.83} & 0.06 \\
    + SynFormer
    & $\sim$97 & 0.44 & 0.88 & 11.7 & 4.9 &-8.0 & 0.88& 3.50 & 0.39 \\
    + sf ($\geq 15 \& \pm 5$)
    & $\sim$59 & 0.47 & 0.91 & 11.3 & 4.5 &-8.3&0.88& 2.00 & 0.35 \\
    + sf ($\geq 15 \&\pm 3$)
    & $\sim$41 & 0.48 & 0.91 & 11.5 & 4.4 &-8.2&0.88& 2.00 & 0.33 \\
    \bottomrule
    \end{tabular}}
    \label{tab:synformer_postprocess}
    \vspace{-18pt}
\end{wraptable}

We evaluate DBMol-Syn to test whether the DBMol signal is preserved after moving molecules into a more synthesizable region of chemical space.
For each DBMol molecule, SynFormer maps it to a structurally similar and synthetically accessible neighbor, without Boltz-2-, AF3-, Vina-, or docking-based filtering.
Thus, DBMol-Syn is a synthesis-aware robustness check rather than an additional score-based selection procedure. For this, we additionally report \textit{Syn.}, the fraction of generated molecules for which AiZynthFinder~\cite{Genheden2020AiZynthFinder} finds a solved retrosynthetic route.

As shown in Table~\ref{tab:synformer_postprocess}, this step substantially improves \textit{syn.}, reaching 0.33 to 0.39 across the three setups. This is close to the strongest baseline, CGFlow-ZS, which achieves a \textit{syn.} score of 0.40, and clearly above all other baselines from the main table, which reach at most 0.27 with Synformer. Meanwhile, SynFormer preserves much of the AF3-based signal.
AF3 Success decreases from $0.92$ to $0.88$, and BCov decreases from $0.51$ to $0.44$.
A simple size-consistency filter recovers much of this gap, reaching $0.91$ AF3 Success and $0.47$ BCov under the $\pm 5$ filter, with improved Dist and iPAE. A stricter filter of $\pm 3$ leads to further improved results on BCov and Dist.
This suggests that the optimized signal is better preserved when the SynFormer neighbor remains close in molecular size, consistent with the fact that DBMol optimizes each molecule under its original size scale.
Overall, synthesis-aware refinement introduces a tradeoff, but much of the DBMol signal transfers to more synthesizable molecules when the mapped neighbor is size-consistent and structurally similar.

\subsection{\textit{De Novo} Molecule Design on Additional Targets}
\begin{wraptable}[13]{r}{0.58\textwidth}
    \centering
    \vspace{-10pt}
    \footnotesize
    \caption{\textbf{De novo molecule design results on additional targets.}
    We compare DBMol with DiffSBDD, a pocket-conditioned structure-based baseline applicable to the strict de novo setting without reference ligands.}
    \setlength{\tabcolsep}{2.8pt}
    \renewcommand{\arraystretch}{1.05}
    \resizebox{1.0\linewidth}{!}{
    \begin{tabular}{@{}lcccccc>{\columncolor{Gray}}c@{}}
    \toprule
    Method 
    & BCov $\uparrow$ 
    & AF3 Succ. $\uparrow$ 
    & iPAE $\downarrow$ 
    & Dist $\downarrow$ 
    & Vina $\downarrow$
    & Div $\uparrow$
    & Avg. Rank $\downarrow$ \\
    \midrule

    \multicolumn{8}{@{}l}{\textit{LGR4}} \\
    DiffSBDD
    & 0.23 & 0.51 & 13.2 & 5.7 & -9.6 & 0.90 & 1.67 \\
    \rowcolor{Gray}
    DBMol
    & 0.25 & 0.71 & 11.4 & 3.9 & -8.1 & 0.87 & \textbf{1.33} \\

    \midrule
    \multicolumn{8}{@{}l}{\textit{CD47}} \\
    DiffSBDD
    & 0.01 & 0.06 & 8.3 & 25.3 & -4.0 & 0.93 & \textbf{1.50} \\
    \rowcolor{Gray}
    DBMol
    & 0.03 & 0.08 & 8.7 & 28.5 & -6.2 & 0.87 & \textbf{1.50} \\

    \bottomrule
    \end{tabular}}
    \label{tab:denovo_targets}
    \vspace{-20pt}
\end{wraptable}

We next evaluate DBMol in a stricter \textit{de novo} setting, which is its intended use case.
Unlike methods requiring reference ligands, pharmacophore annotations, or pocket--ligand training pairs, DBMol only uses the target pocket positions inside the protein sequence, and a structure prediction model at test time. This makes it applicable to new or weakly annotated targets.
Since these proteins do not have a predefined 3D pocket, we define it using the pocket region induced by a high-scoring DBMol-generated molecule, selected by the average of Boltz-2 predicted BCov and ipTM.

We consider two additional targets: LGR4~\cite{wang2025lgr4} and CD47~\cite{zhao2022cd47}.
These provide complementary test cases, including a GPCR target and a protein--protein interaction immune-checkpoint target.
We compare DBMol against DiffSBDD. The results are summarized in \cref{tab:denovo_targets}. 

On LGR4, DBMol improves the AF3-based metrics over DiffSBDD, including BCov, AF3 Success, iPAE, and Dist, while DiffSBDD obtains better Vina and diversity scores. 
On CD47, DBMol improves BCov, AF3 Success, and Vina, while DiffSBDD performs better on iPAE, Dist, and diversity, yielding the same average rank. However, both methods show low absolute pocket engagement, suggesting that CD47 remains challenging in this \textit{de novo} setting.
Overall, DBMol remains competitive on \textit{de novo} targets without using reference ligands or requiring a complete 3D pocket geometry.

\subsection{Additional results}
\paragraph{Optimization--projection tradeoff.}

\begin{figure*}[t]
    \centering
    \begin{subfigure}[b]{0.49\linewidth}
        \centering
        \begin{subfigure}[b]{0.99\linewidth}
            \includegraphics[width=\linewidth]{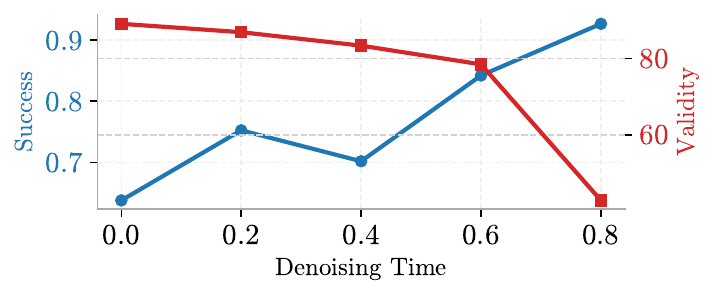}
        \label{fig:start_time}
        \end{subfigure}
    \end{subfigure}
    \hfill
    \begin{subfigure}[b]{0.49\linewidth}
        \centering
        \begin{subfigure}[b]{0.99\linewidth}
            \includegraphics[width=\linewidth]{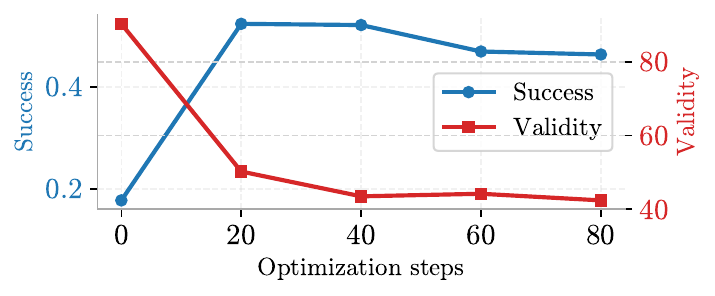}
        \label{fig:boltz_opt}
        \end{subfigure}
    \end{subfigure}
    \vspace{-15pt}
    \caption{Success rate based on Boltz2 and molecular validity as functions of denoising time (left) and optimization steps (right).}
    \label{fig:tradeoff}
\end{figure*}

We analyze the tradeoff between target-specific success and chemical validity.
Stronger structure-predictor guidance can better preserve pocket-specific signals, but it can also move relaxed molecular representations farther from the discrete manifold of valid molecule graphs.
We vary two control parameters: the denoising initialization time and the number of Boltz-guided optimization steps.

As shown in the left of \cref{fig:tradeoff}, initializing denoising at later diffusion times makes the final generated molecules closer to the optimized relaxed representation.
This improves the preservation of Boltz-guided structural signals and leads to higher success rates.
However, because fewer denoising steps remain, the model has less opportunity to correct chemically invalid structures, resulting in a clear decrease in chemical validity.

The right panel of \cref{fig:tradeoff} isolates the effect of Boltz-guided gradient refinement by evaluating at different optimization steps, while keeping the projection and evaluation protocol fixed. 
Proxy-based metrics (Boltz2 Success) improve rapidly in the early stages and then saturate, indicating that the gradient updates directly contribute to the observed improvement rather than merely increasing the candidate pool. 
At the same time, validity decreases with excessive optimization. 

\paragraph{Additional analyses.}We provide additional analyses in the appendix.
In App.~\ref{app:condition}, we show that DBMol supports DBMol-candidate-conditioned iterative refinement by further optimizing a promising molecule found by DBMol. In App.~\ref{app:sim}, we show it finds molecules close to known actives, without directly copying known ligands.
In App.~\ref{app:defog}, we discuss the choice of the denoiser.
In App.~\ref{app:ablations}, we study the necessity of using both contact- and affinity-related losses.

\section{Conclusion}
\label{sec:conclusion}

We introduced DBMol, an optimization-centric framework that leverages differentiable signals from modern protein--ligand structure predictors for small-molecule design without docking supervision, reference ligands, or curated protein--ligand datasets.
Results with held-out metrics on LIT-PCBA show that structure-predictor-guided optimization improves pocket coverage while maintaining molecular diversity over unconditional generation, and stays competitive with other baselines, supporting the practical utility of structure-based guidance.

Several directions remain open and motivate future work.
First, DBMol relies on surrogate objectives, which are inevitably imperfect and may not always align with experimental binding.
As structure prediction models improve, DBMol may benefit from more accurate and better-aligned optimization signals.
A key current limitation is the scarcity of high-quality protein--ligand binding datasets~\cite{vskrinjar2025have}; larger and more open datasets, such as OpenBind~\cite{openbind}, could provide more robust evaluation signals closer to real-world design objectives.
In addition, extending DBMol from 2D molecular graphs to full 3D protein--ligand representations would enable richer geometric reasoning and tighter integration with structure prediction models.
Finally, improving computational efficiency, particularly through parallelization and reduced overall cost, remains an important direction. We also aim to support controllable molecule size during optimization stage.

Overall, DBMol substantially improves over unconditional generation under Boltz2-based metrics, while remaining competitive with reinforcement-learning-based and ligand- or interaction-conditioned baselines under held-out structural evaluation.
These results suggest that structure prediction models can serve as useful differentiable surrogate objectives for target-specific molecular generation, while experimental binding validation remains an important direction for future work.

\section*{Acknowledgements}
This work was funded by ELSA – European Lighthouse on Secure and Safe
AI funded by the European Union under Grant Agreement No. 101070617. This work was supported by the Medical Research Council as part of UK Research and Innovation (UKRI) (MC\_UP\_A025\_1013 to S.H.W.S.). 
Y.Q. was supported by the Swiss National Science Foundation (SNSF grant 10001445).

\bibliographystyle{unsrt}
\bibliography{tools/clean_ref_defog}

\appendix
\clearpage
\newpage
\onecolumn
\addcontentsline{toc}{section}{Appendix} 
\part{Appendix} 
\parttoc      
\section*{Impact Statement}
\label{app:impact}

This paper introduces DBMol, a computational framework for structure-guided molecular design that leverages differentiable signals from protein--ligand structure prediction models.
The proposed method is intended to advance methodological research in molecular generation and structure-based drug design.
While structure-guided molecular design may have downstream applications in biomedical research, this work focuses on algorithmic development and evaluation.
We do not anticipate any immediate societal or ethical impacts arising from the proposed methodology.

\section{Discussions}

\subsection{Alternatives}

A possible alternative formulation of our problem is conditional generation. Since Boltz-2 provides differentiable gradients in the logits space, these gradients could, in principle, be used to modulate the rate matrix of the flow matching denoising process. This would inject structural guidance from the structure prediction model directly during sampling.

In practice, however, this approach introduces substantial computational and optimization challenges. First, it requires calling Boltz-2 to compute gradients at each denoising step. A single backward pass of Boltz-2 takes approximately 30 seconds. Even if the conditional signal is applied only at a small number of time steps, the total cost remains prohibitively high. Second, when the time variable $t$ is close to 0, the conditional gradients are dominated by noise and become ineffective. In contrast, near the end of sampling, the logits are highly peaked and approach one-hot vectors. Meaningful modification at this stage requires a very large condition weight, which leads to unstable optimization and makes the process difficult to tune.

Moreover, even after applying the alignment procedure between Boltz2 and DeFoG described in App.~\ref{app:boltz_setup} to align Boltz-2 gradients, additional structural difficulties arise. The molecular graphs used by Boltz-2 and DeFoG differ in node ordering and construction. As a result, gradients must be carefully aligned through typed graph isomorphism before they can be applied. This alignment requires simultaneously constraining node types and edge types to prevent ambiguous matches caused by purely topological isomorphisms. Although we implemented this alignment pipeline, the isomorphism step remains difficult to stabilize for certain cases, particularly involving aromatic bonds, further increasing system complexity.

For these reasons, we adopt this two-stage approach proposed in this work, which decouples structural guidance from molecular generation at the modeling level. This design avoids the high cost of step-wise guidance, the instability and tuning difficulty of conditional denoising, and the implementation complexity introduced by cross-representation graph isomorphism. As a result, the overall pipeline is significantly more efficient and robust.

\section{Background}
\label{app:background}

\subsection{Discrete Flow Matching}
\label{app:dfm}

In generative modeling, the primary goal is to generate new data samples from the underlying distribution that produced the original data, $\bm{p}_\mathrm{data}$.
An effective approach is to learn a mapping between a simpler distribution $\bm{p}_\epsilon$ that can be easily sampled, and $\bm{p}_\mathrm{data}$.

We consider a discrete variable $z_t \in \mathcal{Z} = \{1, \ldots, Z\}$ evolving over time $t \in [0,1]$.
Its marginal distribution is denoted by $\bm{p}_t \in \Delta^Z$.
The initial distribution is set to a predefined noise distribution, $\bm{p}_0 = \bm{p}_\epsilon$, while $\bm{p}_1 = \bm{p}_\text{data}$ represents the target data distribution.
We refer to the mapping $t:1 \to 0$ as the noising process and $t:0 \to 1$ as the denoising process.

Following \cite{campbell2024generative}, the noising trajectory is defined through a simple linear interpolation conditioned on a datapoint $z_1$:
\begin{equation}
    p_{t|1}(z_t|z_1) = t\,\delta(z_t, z_1) + (1-t)\,p_0(z_t).
\end{equation}
A common choice for $p_0$ is the uniform distribution over $\mathcal{Z}$.

\paragraph{Denoising as a CTMC.}
The denoising process is formulated as a continuous-time Markov chain (CTMC).
A CTMC is characterized by a rate matrix $\bm{R}_t \in \mathbb{R}^{Z \times Z}$, which governs the instantaneous transition rates between states.
For an infinitesimal time step $\mathrm{d}t$, the transition probabilities satisfy:
\begin{equation}\label{eq:dfm_sampling}
    p_{t+\mathrm{d}t|t}(z_{t+\mathrm{d}t}|z_t)
    = \delta(z_t, z_{t+\mathrm{d}t}) + R_t(z_t, z_{t+\mathrm{d}t})\,\mathrm{d}t .
\end{equation}

By definition, $R_t(z_t, z_{t+\mathrm{d}t}) \ge 0$ for $z_t \neq z_{t+\mathrm{d}t}$, and the diagonal entries are set to ensure normalization.
The evolution of the marginal distribution follows the Kolmogorov forward equation,
$\partial_t \bm{p}_t = \bm{R}_t^\top \bm{p}_t$.
\paragraph{Conditional rate construction.}
Similarly to the noising process, denoising is performed under conditioning on $z_1$.
We consider a $z_1$-conditional rate matrix $\bm{R}_t(\cdot,\cdot|z_1)$ that satisfies the corresponding Kolmogorov equation.
Under mild assumptions, \cite{campbell2024generative} derive a closed-form expression for a valid conditional rate matrix.
For $z_t \neq z_{t+\mathrm{d}t}$, it is given by:
\begin{equation}
    R^*_t(z_t, z_{t+\mathrm{d}t}|z_1)
    = \frac{\operatorname{ReLU}[\partial_t p_{t|1}(z_{t+\mathrm{d}t}|z_1)
    - \partial_t p_{t|1}(z_t|z_1)]}{Z^{>0}_t\,p_{t|1}(z_t|z_1)}.
\end{equation}

Intuitively, this rate redistributes probability mass toward states that require increased marginal probability.
The unconditional rate matrix used for sampling is obtained by marginalizing over $z_1$:
\[
R_t(z_t, z_{t+\mathrm{d}t})
= \mathbb{E}_{p_{1|t}(z_1|z_t)}[R_t(z_{t+\mathrm{d}t}, z_t|z_1)].
\]
\subsection{Discrete Flow Matching for Graphs}
\paragraph{Graph notation.}
We denote by $G = (x^{(1:n:N)}, c^{(1:n:N)}, e^{(1:i \neq j:N)})$ a directed graph with $N$ nodes.
Nodes, charges, and edges are categorical variables with $x^{(n)} \in \{1,\ldots,X\}$, $c^{(n)} \in \{1,\ldots,C\}$, and $e^{(i,j)} \in \{1,\ldots,E\}$.

\paragraph{Training objective.}
The denoising model is parameterized by a neural network with parameters $\theta$, trained to predict the clean node, charge, and edge distributions given a noisy graph $G_t$.
The predicted distributions are denoted by $\bm{p}^\theta_{1|t}(\cdot|G_t)$.

Training is performed using a cross-entropy loss applied independently to nodes, charges, and edges:
\begin{equation}
    \mathcal{L} = \mathbb{E}_{t,G_1,G_t}\!\left[
    -\sum_n \log p_{1|t}^{\theta,(n)}(x_1^{(n)}|G_t)
    - \lambda_c \sum_n \log p_{1|t}^{\theta,(c,n)}(c_1^{(n)}|G_t)
    - \lambda_e \sum_{i \neq j} \log p_{1|t}^{\theta,(i,j)}(e_1^{(i,j)}|G_t)
    \right].
\end{equation}

\paragraph{Denoising and sampling.}
At sampling time, denoising is formulated as a CTMC evolving from an initial graph $G_0 \sim \bm{p}_0$ toward $t=1$.
The transition kernel is defined by a rate matrix $\bm{R}_t$, approximated using the network predictions $\bm{p}^\theta_{1|t}$.
In practice, the continuous-time dynamics are discretized using a finite step size $\Delta t$, resulting in an Euler-style update.

\paragraph{Algorithmic summary.}
For clarity, the overall training and sampling procedures of DeFoG are summarized in \cref{alg:training} and \cref{alg:sampling}, respectively.

\begin{figure*}[h]
    \centering
    \begin{minipage}[t]{0.49\textwidth}
        \centering
        \begin{algorithm}[H]
        \caption{DeFoG Training}
            \begin{algorithmic}[1]
                \State \textbf{Input:} Graph dataset $\mathcal{D} = \{ G^1, \ldots, G^M\}$
                \While{$f_\theta$ not converged}
                    \State Sample $G \sim \mathcal{D}$
                    \State Sample $t \sim \mathcal{T}$
                    \State Sample $G_t \sim p_{t|1}(G_t | G)$ \Comment{\ca{Noising}}
                    \State $h \gets \operatorname{RRWP}(G_t)$ \Comment{\ca{Extra features}}
                    \State $\bm{p}^{\theta}_{1|t}(\cdot|G_t) \gets f_\theta(G_t, h, t)$  \Comment{\ca{Denoising prediction}}
                    \State $\operatorname{loss} \gets 
                    \operatorname{CE}_\lambda (G, \,\bm{p}^{\theta}_{1|t}(\cdot|G_t) )$ 
                    \State $\operatorname{optimizer.step}(\operatorname{loss})$
                \EndWhile
            \end{algorithmic}
        \label{alg:training}
        \end{algorithm}
    \end{minipage}
    \hfill
    \begin{minipage}[t]{0.49\textwidth}
        \centering
        \begin{algorithm}[H]
        \caption{DeFoG Sampling}
        \begin{spacing}{1.04}
            \begin{algorithmic}[1]
                \State \textbf{Input:} $\#$ graphs to sample $S$
                \For{$i = 1$ \textbf{to} $S$}
                    \State Sample $N$ from train set  \Comment{\ca{\# Nodes}}
                    \State Sample $G_0 \sim p_0 (G_0)$
                    \For{$t = 0$ \textbf{to} $1- \Delta t$ \textbf{with step} $\Delta t$}
                        \State $h \gets \operatorname{RRWP}(G_t)$ \Comment{\ca{Extra features}}
                        \State $\bm{p}^{\theta}_{1|t}(\cdot|G_t)\gets f_\theta(G_t, h, t)$ \Comment{\ca{Denoising prediction}}
                        \State $G_{t+\Delta t} \sim \Tilde{p}_{t + \Delta t | t} (G_{t + \Delta t}| G_t )$\Comment{\ca{\Cref{eq:dfm_sampling}}}
                    \EndFor
                    \State Store $G_1$
                \EndFor
            \end{algorithmic}
        \end{spacing}
        \label{alg:sampling}
        \end{algorithm}
    \end{minipage}
    \vspace{-12pt}
\end{figure*}

\paragraph{Discussion.}
Discrete flow matching offers an efficient and flexible framework for generative modeling in discrete domains. A key advantage of this formulation is the decoupling between training and sampling: the denoising model is trained to predict clean data distributions, while the sampling dynamics are specified separately through a continuous-time stochastic process. This separation is particularly advantageous for graph generation, where rigid sampling schemes can be computationally expensive due to the combinatorial structure of nodes and edges. As demonstrated in DeFoG~\cite{qin2025defog}, this flexibility can be leveraged through tailored sampling strategies, including modified rate constructions and adaptive denoising schedules, to substantially reduce the number of generated steps required at inference time without degrading generation quality. These properties make discrete flow matching especially suited for scalable graph generation and motivate its use as the foundation of our approach.

\section{Experimental Details}\label{app:experimental_details}

This section provides further details on the experimental settings used in the paper.



\subsection{Boltz2 Gradient Computation Setup.}
\label{app:boltz_setup}

This section describes how gradients are extracted from the Boltz2 model~\cite{passaro2025boltz} and incorporated into the DeFoG sampling procedure. The design goal is to use Boltz2 as an external differentiable evaluator that provides guidance signals. We therefore focus on minimal and targeted modifications to the Boltz2 inference pipeline and on a clean interface for mapping gradients into discrete sampling.

\paragraph{Enabling gradient support in Boltz2.}
To enable gradient-based guidance, Boltz2 must expose gradients with respect to molecule structure variables while preserving stable structure prediction and affinity estimation. The modifications described below are restricted to inference-time behavior and do not change the trained model parameters.

\paragraph{Differentiable relaxation of discrete inputs.}
Boltz-2 takes SMILES strings as input and preprocesses them into a structured object, denoted as \texttt{batch}, which contains all molecular features required for inference.
Among these features, we focus on three components that are relevant for gradient-based optimization: atom types, bond types, and atomic charges.

First, Boltz-2 represents atom types using a large atomic vocabulary, where each atom is encoded by an index corresponding to its chemical element, for example carbon is indexed as 6, stored as \texttt{batch["ref\_element"]}
In contrast, DeFoG operates on a much smaller, fixed set of atom-type categories, where carbon corresponds to a single category in the DeFoG vocabulary.
This discrepancy prevents atom-type representations from Boltz-2 from being updated directly in the DeFoG space.
To resolve this mismatch, we first map Boltz-2 atom types to the subset of atom types supported by DeFoG, and restrict gradient updates to these matched categories.
We then treat atom types as continuous variables by converting their discrete one-hot representations into continuous categorical vectors, which enables gradient-based optimization while keeping unsupported atom types fixed.

Bond types in Boltz-2 are encoded as integers and are passed through an embedding layer. To make this embedding step differentiable, we explicitly construct one-hot bond-type representations instead of directly encoding the type index integer, stored as \texttt{batch["type\_bonds\_onehot"]}.
We then modify the implementation inside the Boltz-2 model so that the bond embedding is applied via matrix multiplication, which allows gradients to propagate through the bond-type representation during optimization.

Last, atomic charges are modeled as continuous scalar values throughout the pipeline, stored as \texttt{batch["ref\_charge"]}. Since this term is concatenated into the atom feature tensor, it does not naturally align with categorical representations; we therefore keep it in its original scalar form. As a result, gradients with respect to atomic charges can be computed directly, without requiring any additional relaxation or approximation.

\paragraph{Losses.}
\texttt{Boltz-2} confidence outputs include several complementary estimates of complex quality, and their values may depend on recycling convergence and low-confidence or disordered regions. 
We therefore define $\mathcal{L}_{\mathrm{conf}}$ as a weighted combination of selected confidence terms from the Boltz-2 confidence model output. 
Specifically, we use terms such as \texttt{complex\_iplddt}, which encourages high local confidence around interface tokens; \texttt{complex\_ipde} and \texttt{complex\_pde}, which penalize large predicted distance errors at the interface and over the complex; \texttt{ptm}, which encourages reliable global complex geometry; and \texttt{confidence\_score}, an aggregate confidence measure. 
For scores where higher values indicate higher confidence, we use their negative values as losses, while for error-based quantities such as PDE/ipDE, we minimize the predicted error directly. 
The exact weighted composition of $\mathcal{L}_{\mathrm{conf}}$ is provided in App.~\ref{app:experimental_details}.

Boltz2 also predicts protein-ligand binding \texttt{affinity}, which reflects how strongly a small molecule binds to a target protein and provides a direct optimization signal for ligand generation. The affinity model outputs scalar predictions through \texttt{affinity\_pred\_value}. We use this value directly as the loss function and minimize it to bias generation toward higher-affinity binding candidates. At the same time, we optimize the binder/non-binder classification signal \texttt{affinity\_probability\_binary}. Specifically, we maximize in logit space \texttt{affinity\_logits\_binary}, whose sigmoid corresponds to the predicted probability that the ligand binds to the target. Both terms are equally weighted in our loss function.

We define a \texttt{contact loss} based on Boltz2-predicted distance distributions (\texttt{p-distogram}) to encourage protein-ligand contacts. Details are provided in \cref{sec:boltz}.


\paragraph{Memory-efficient gradient computation.}
Direct gradient computation in Boltz2 is memory-intensive. We apply three optimizations. First, the \texttt{recycling} mechanism is limited to a single refinement step, which substantially reduces memory usage while preserving gradient quality in practice. Second, gradient checkpointing is applied to pairwise representation update modules for lower memory consumption. Third, the gradient does not pass through the structure module of Boltz-2.

\paragraph{Requirement computational resources and running time.} GPU memory usage for Boltz-2–based optimization primarily depends on protein length. In our experiments, this step consistently requires less than 65 GB of GPU memory.
For molecule generation and evaluation, all setups require less than 20 GB of GPU memory. All experiments are performed on NVIDIA A100, H100, or H200 GPUs.
For efficiency, MSA searches are performed once per protein and cached. During optimization, we reuse these preprocessed MSAs since the protein sequence stays unchanged, resulting in an average runtime of approximately 30 seconds per optimization step.

\paragraph{Hyperparameters.}
In our framework, the molecular size is fixed at the beginning of generation. While this size could be inferred heuristically from the pocket size or searched over in parallel, we use the reference-ligand size for targets with a known ligand to keep the pipeline simple.
This is the only use of reference-ligand information in our pipeline; no reference-ligand geometry, chemical identity, or interaction pattern is used during optimization.

For the main LIT-PCBA benchmark and de novo design results, we use a single default optimization configuration across all proteins. 
In particular, we fix the optimization hyperparameters and report results after 50 optimization iterations for every target, rather than selecting target-specific iteration counts. 
In practice, we use SGD with learning rate $\eta=2.0$ for relaxed molecule optimization. 
The loss weights are fixed across all LIT-PCBA experiments, with  $w_1=2.5$, and $w_2=2.5$. 
For the contact loss, we use $d_{\mathrm{cutoff}}=8.0$~\AA, and assign equal weights to the pocket-averaged term and the anchoring term. 
The confidence loss $\mathcal{L}_{\mathrm{conf}}$ combines the nonzero-weight Boltz-2 confidence terms, including ipTM, ipLDDT, ipDE, PDE, confidence score, and pTM, with weights $1.0$, $0.3$, $0.3$, $0.1$, $0.3$, and $0.1$, respectively. 
All other confidence terms are assigned zero weight. 
For the projection stage, we initialize the discrete denoising process at $t_0=0.8$. 
The same default configuration is used for the component ablations unless otherwise stated.

All generated molecules are evaluated with the same AF3-based evaluation pipeline. 
AF3 metrics are not used during molecular optimization; they are used after generation for evaluation and target-level reporting. 

Due to the computational cost of AF3-based evaluation, we perform only a limited hyperparameter search over the loss terms rather than exhaustive tuning. 
The reported results should therefore be interpreted as performance under a practical, modest tuning budget. 
More systematic hyperparameter optimization may further improve performance, especially for additional de novo targets and target-specific design settings; we leave this direction to future work.

\subsection{Mapping Boltz2 Gradients to DeFoG Sampling}

\paragraph{Handling aromatic bond representations.}
Boltz2 and DeFoG may adopt different conventions for aromatic bonds, such as explicit aromatic categories or kekulized representations. To avoid inconsistencies during alignment and gradient transfer, we use DeFoG checkpoints trained with explicit aromatic bond types. This choice simplifies matching and removes the need to redistribute gradient mass across bond categories.

\paragraph{Projection of gradients into the DeFoG state space.}
Gradients are projected into the discrete state spaces supported by DeFoG. Atom-type gradients are mapped from Boltz’s larger atomic vocabulary to the element set used by DeFoG, namely \{\texttt{C}, \texttt{N}, \texttt{O}, \texttt{F}, \texttt{P}, \texttt{S}, \texttt{Cl}, \texttt{Br}, \texttt{I}\}. Bond-type gradients are mapped to DeFoG edge categories \{\texttt{NO-EDGE}, \texttt{SINGLE}, \texttt{DOUBLE}, \texttt{TRIPLE}, \texttt{AROMATIC}\}. Charge gradients are compressed into a discrete set of charge states $\{0, +1, -1\}$.

\subsection{DeFoG Setup}
\label{app:defog_setup}

\paragraph{Denoising neural architecture.}
DeFoG's denoising neural network takes a noisy graph $G_t$ as input and predicts the clean marginal probability for each node $x^{(n)}$ via $p^{\theta,(n)}_{1|t} (\cdot|G_t)$ and for each edge $e^{(ij)}$ via $p^{\theta,(ij)}_{1|t} (\cdot|G_t)$.  This formulation boils down the graph generative task to a graph-to-graph mapping.
While both message-passing layers and graph transformers can be used for this task, graph transformers have empirically outperformed message-passing layers in graph generation~\citep{qin2023sparse}.
DeFoG thus adopts the transformer architecture of~\cite{vignac2022digress}, using multi-head attention layers which encode node, edge, and graph-level features while preserving node permutation equivariance.

\paragraph{Enhancing model expressivity.}
Graph neural networks, including graph transformers, have inherent limitations in their expressive power~\citep{xu2018powerful,zhu2023structural}. An usual approach to overcome the limited representation power of graph neural networks consists of explicitly augmenting the inputs with features that the networks would otherwise struggle to learn.
We adopt Relative Random Walk Probabilities (RRWP) encodings that are proved to be expressive for both discriminative~\citep{ma2023graph} and generative settings~\citep{siraudin2024cometh}.
RRWP encodes the likelihood of traversing from one node to another in a graph through random walks of varying lengths, offering insights into graph dynamics across different hop distances.
In particular, given a graph with an adjacency matrix $A$, we generate $K-1$ powers of its degree-normalized adjacency matrix, $M=D^{-1}A$, i.e., $\left[I, M, M^2, \ldots, M^{K-1}\right]$. 
We concatenate the diagonal entries of each power to their corresponding node embedding, while combining and appending the non-diagonal to their corresponding edge embeddings.

\paragraph{Training setup.} We encode aromatic bonds as an additional bond type and explicitly model formal charges during training. The model is trained on a single NVIDIA A100 GPU for approximately 50 hours for 300 epochs. Model checkpoints are selected based on molecule validity evaluated on a validation set, with the final checkpoint chosen after 38 hours of training.

\paragraph{Sampling setup.} We use the \texttt{polydec} time distortion function in DeFoG for both training and sampling to improve efficiency. During sampling, the denoising process is run for 200 steps with an initial time $t_0 = 0.8$. We find that $t_0 = 0.8$ provides a good trade-off between generating valid molecules and preserving the strength of Boltz-derived optimization signals. Reducing the number of denoising steps leads to a noticeable drop in validity. We therefore use 200 steps as a balanced choice that maintains both sample quality and computational efficiency.
For each experiment, we generate 2048 molecules and randomly select 100 valid molecules for evaluation.
We adopt this oversampling strategy because the guided sampling distribution can deviate substantially from the clean data distribution on which DeFoG is trained, leading to a non-negligible fraction of invalid samples. Despite this, the overall generation process remains highly efficient: sampling involves only a small number of denoising steps over relatively small molecule graphs, resulting in a total runtime of under 10 seconds (8 seconds for generation, 2 seconds for filtering) for all generated samples.

\subsection{Dataset Details}\label{app:datasets_details}
\paragraph{LIT-PCBA.}
LIT-PCBA~\cite{litpcba} is a carefully curated virtual screening benchmark constructed from 149 dose-response PubChem BioAssays. The dataset is filtered to remove false positives, assay artifacts, and major physicochemical imbalances, and retains only targets with experimentally resolved ligand-bound structures. After target selection and asymmetric validation embedding (AVE), LIT-PCBA consists of 15 targets with 7,844 confirmed actives and 407,381 confirmed inactives.
We fix the size of the molecules during generation based on the reference ligand, which is considered as the only information leakage in our generation pipeline.
We also report results for molecules with a fixed size of 25 atoms per protein, corresponding to the median reference molecule size across the 7 proteins, and use this setting for ablation studies as shown in App.~\ref{app:ablations}.
In our experiments, we evaluate on 7 targets: Estrogen Receptor (ESR; PDB IDs: 2iok and 2p15), Glucocerebrosidase (GBA; 2v3d), Mitogen-Activated Protein Kinase 1 (MAPK1; 4zzn), Aldehyde Dehydrogenase 1 (ALDH1; 5l2m), Tumor Protein p53 (TP53; 3zme), and Vitamin D Receptor (VDR; 3a2i).

\paragraph{ZINC-250K.}
ZINC-250K is a curated subset of the ZINC database~\cite{sterling2015zinc} containing approximately 250,000 structurally diverse, drug-like molecules. It is widely used in molecule machine learning for tasks such as property prediction and de novo molecule generation.

\subsection{Synformer Setup}\label{app:synformer}

For the synthesis-aware variant in \cref{tab:synformer_postprocess}, we apply SynFormer to all 100 valid DBMol molecules per target without any pre-filtering based on Boltz-2, AF3, Vina, or docking scores. For SynFormer expansion, we use fixed hyperparameters across all experiments: exhaustiveness=128, time-limit=300, search-width=32, and num-smiles-per-target=100. For the metrics \textit{syn.}, we run AiZynthFinder~\cite{Genheden2020AiZynthFinder} with default settings.

We additionally report results under simple molecule-size consistency filters. The '+size filter (+-N)' rows in \cref{tab:synformer_postprocess} retain only SynFormer outputs whose heavy-atom count differs from the input DBMol molecule by at most N atoms. This filter is based solely on molecular size, not on any evaluation metric.

\subsection{Baseline Setup}\label{app:baseline}
We mainly follow \cite{rekesh2025syncogen} setup for our baselines. Precisely,

\begin{itemize}
    \item SynCoGen. SynCoGen~\cite{rekesh2025syncogen} is an amortized molecular generator that conditions directly on pharmacophore profiles, specified as interaction types and their 3D positions. During training, pharmacophore features are provided as input to the model; to encourage robustness and generalization, at most seven pharmacophore features are randomly subsampled per example. At inference time, SynCoGen generates molecules conditioned solely on the provided pharmacophore profile, without requiring reference ligands themselves.
    \item ShEPhERD. ShEPhERD~\cite{adamsShEPhERD2024} is a 3D molecular generator conditioned on pharmacophore interaction profiles.
    While ShEPhERD explicitly models spatial interactions, it does not impose synthesizability constraints, and chemical feasibility is not guaranteed by design.
    \item SynFormer.
    SynFormer~\cite{gao2024generative} is a synthesis-aware 2D molecular generator that produces compounds by modeling chemically valid reaction pathways.
    It generates molecules by conditioning on a reference compound and optimizing structural similarity under 2D fingerprint-based metrics.
    \item CGFlow-ZS.
    CGFlow~\cite{shen2025compositional} is a pathway-based generative model that produces molecules together with 3D poses by sampling synthesis pathways.
    In its original formulation, CGFlow relies on reinforcement learning with pocket-conditioned rewards.
    To align with amortized sampling, we use a zero-shot variant (CGFlow-ZS) that samples using the pocket-conditional generation proposed by ~\cite{shen2024tacogfn}, without task-specific fine-tuning. CGFlow-ZS results in \cref{tab:main_results} are averaged over 6 of the seven LIT-PCBA targets without 3a2i due to missing results from baselines.
    \item DiffSBDD~\cite{schneuing2024structurebaseddrugdesignequivariant}. DiffSBDD is a standard 3D pocket-conditioned model for structure-based molecule generation.  It can generate molecules without requiring a known reference ligand, but it requires a well-defined binding pocket.  For instance, an overly small pocket can naturally bias the model toward generating small molecules.  In our experiments, for each target, we define the pocket as the residues within $8~\mathrm{\AA}$ of the Boltz-2-predicted ligand-binding region. For de novo design, to have a proper definition for the 3D pocket, we define it using the pocket region induced by a high-scoring DBMol-generated molecule.
\end{itemize}

\subsection{Evaluation Details}\label{app:eval}

\subsubsection{AF3-based structural evaluation}
We evaluate each generated protein--ligand complex with AlphaFold3 (AF3).
For each target protein, we provide the corresponding multiple sequence alignment (MSA).
For each ligand, we run AF3 once with a fixed random seed and generate one diffusion sample.

\paragraph{Confidence and interface quality.}
AF3 reports the inter-chain predicted TM score (ipTM), which measures confidence in the predicted protein--ligand interface.
We report iPAE, the predicted interface aligned error, where lower values indicate higher predicted interface quality. 
We also use ipTM $\geq 0.6$ as the high-confidence criterion, as discussed for AF3 Success.

\paragraph{Pocket engagement.}
To evaluate whether the ligand is placed in the intended binding pocket, we use the target pocket residues.
For each pocket residue, we compute the minimum heavy-atom distance to the ligand.
We report BCov, defined as the fraction of pocket residues within 3.5~\AA\ of any ligand atom, and Dist, defined as the mean of these per-residue minimum distances.
Higher BCov indicates broader pocket coverage, while lower Dist indicates more localized pocket binding.

\paragraph{AF3 Success.}
We define AF3 Success as a heuristic thresholded structural metric built from AF3 outputs.
A predicted complex is counted as successful if it satisfies both ipTM $\geq 0.6$ and at least one target pocket residue (defined based on the reference ligand-protein structure of LIT-PCBA, and chosen manually for de novo targets) lies within 3.5~\AA\ of the ligand.
The threshold was selected without considering DBMol or its variants: we evaluated several ipTM thresholds on unconditional generation and other baselines, and chose this criterion because it provided a basic separation between unconditional generation and stronger methods.
Since thresholded metrics can affect rankings, our conclusions do not rely on AF3 Success alone, but on the full set of AF3-based metrics, including BCov, iPAE, and Dist.

\subsubsection{Other evaluation metrics}

\paragraph{Boltz2-aligned metrics.}
We report Boltz2 Success as a proxy-aligned reference metric.
A molecule is counted as Boltz2-successful if it achieves predicted affinity below 0 and target pocket coverage ratio above 0.5 under Boltz-2.
Because DBMol is optimized using Boltz-based signals, Boltz2 Success is reported as a reference metric rather than used as the primary basis for comparison.

\paragraph{Molecular diversity.}
We measure molecular diversity using ECFP4 fingerprints.
For each generated set, we compute the average pairwise Tanimoto distance between molecules, with higher values indicating greater chemical diversity.

\paragraph{Docking-based reference.}
We report AutoDock Vina scores as an additional held-out docking-based reference metric.
For each generated molecule, the receptor and ligand are converted to PDBQT format, and docking is performed in a local search box centered on the Boltz-predicted ligand region with fixed padding.
We use \texttt{exhaustiveness}=32 and \texttt{num\_modes}=10, and report the best-scoring pose among the returned modes.
Lower Vina scores indicate stronger predicted docking affinity.
Vina is used only for evaluation and does not participate in DBMol optimization, candidate selection, or hyperparameter tuning.

\paragraph{Known-active similarity and interaction recovery.}
For LIT-PCBA targets with known active compounds, we additionally evaluate whether generated molecules recover active-like chemical and interaction patterns.
We report similarity to the crystallized reference ligand, active enrichment EF@1\%, and interaction fingerprint similarity (IFP Sim) to the reference ligand.
These metrics help assess whether generated molecules copy known binders or instead discover alternative active-like candidates with related pocket-level interactions.

\section{Additional Results}
\label{app:additional_res}

\subsection{Ablations}
\label{app:ablations}

\begin{figure*}[ht]
    \centering
    \begin{subfigure}[b]{0.99\linewidth}
            \includegraphics[width=\linewidth]{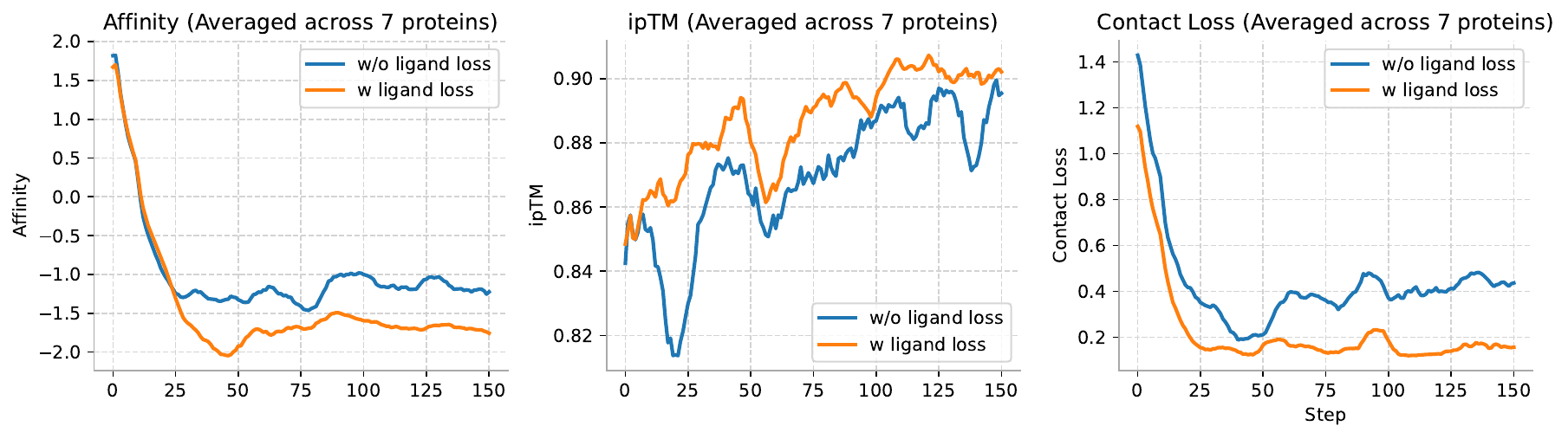}
    \end{subfigure}
    \caption{
    Ligand-centric anchoring loss helps to stabilize the optimization.
    } 
    \label{fig:anchor}
\end{figure*}

\paragraph{Anchoring loss.}We show in \cref{fig:anchor} the effectiveness of the anchoring loss.
Intuitively, the anchoring loss simplifies the optimization by avoiding the need to minimize contact loss independently at all pocket positions, and instead encourages a global attraction between the ligand and the protein.
This stabilizes the optimization process and leads to consistent improvements across all three objectives.

\begin{figure*}[ht]
    \centering
    \begin{subfigure}[b]{0.99\linewidth}
        \centering
        \begin{subfigure}[b]{0.99\linewidth}
            \includegraphics[width=\linewidth]{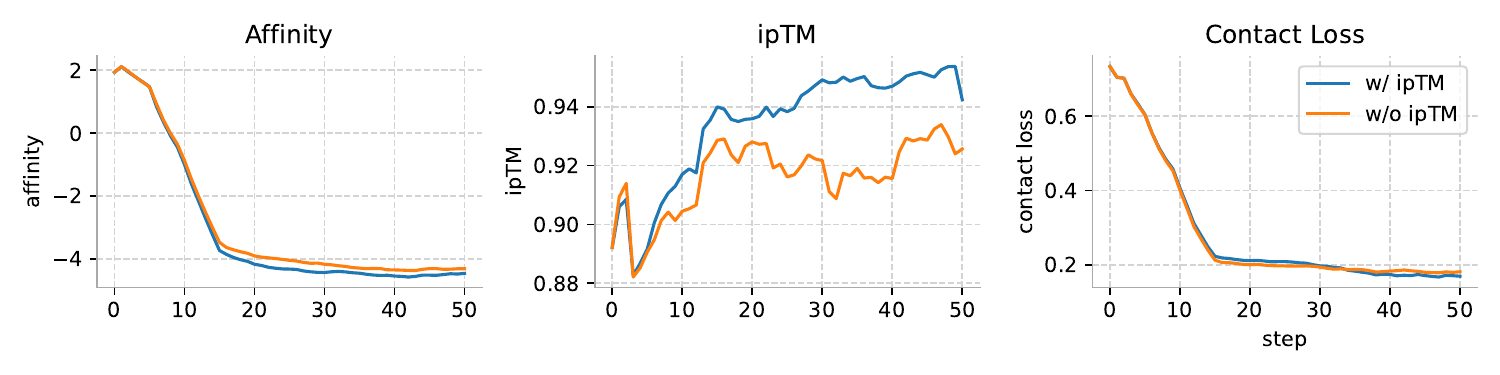}
        \end{subfigure}
    \end{subfigure}
    \caption{Effect of ipTM on Estrogen Receptor (ESR; 2p15)}
    \label{fig:iptm}
    \label{fig:ablations}
\end{figure*}

\paragraph{Effect of ipTM loss.} As shown in \cref{fig:iptm}, removing the ipTM loss does not noticeably affect either affinity or contact loss, while ipTM scores drop significantly across most optimization iterations. This shows the effectiveness of the optimization and also shows that improving the confidence does not sacrifice the other two objectives.

\begin{table}[ht]
    \centering
    \small
    \caption{\textbf{Ablation of affinity and contact objectives.}
    We compare the full DBMol objective with variants that remove either the contact term or the affinity term.
    Boltz2 Success is reported as a proxy-aligned reference metric, while AF3-based metrics provide held-out structural evaluation.}
    \setlength{\tabcolsep}{3.5pt}
    \renewcommand{\arraystretch}{1.05}
    \begin{tabular}{@{}lccccccc@{}}
    \toprule
    Method 
    & BCov $\uparrow$ 
    & AF3 Succ. $\uparrow$ 
    & iPAE $\downarrow$ 
    & Dist $\downarrow$ 
    & Div. $\uparrow$ 
    & Vina $\downarrow$ 
    & Boltz2 Succ. $\uparrow$ \\
    \midrule
    DBMol 
    & 0.54 & 0.97 & 12.0 & 4.1 & 0.89 & -7.6 & 0.47 \\
    DBMol w/o contact 
    & 0.54 & 0.94 & 12.2 & 3.9 & 0.89 & -7.2 & 0.53 \\
    DBMol w/o affinity 
    & 0.47 & 0.90 & 12.8 & 4.7 & 0.90 & -7.9 & 0.80 \\
    \bottomrule
    \end{tabular}
    \label{tab:loss_ablation}
\end{table}

\paragraph{Effect of affinity and contact objectives.}
We ablate the two main Boltz-guided objective terms in \cref{tab:loss_ablation} on 2v3d, 2p15, and 3zme.
The full DBMol objective achieves the best overall AF3-based performance, with the highest AF3 Success and the lowest iPAE among the three variants, while maintaining competitive BCov and Dist.
Removing the contact term slightly improves Dist and Boltz2 Success, and removing the affinity term further increases Boltz2 Success, but both ablations degrade the held-out AF3-based metrics, especially AF3 Success and iPAE.
This indicates that the affinity and contact terms are complementary: optimizing a single proxy term can improve some aligned scores, but the full objective provides a more balanced structural signal under held-out AF3 evaluation.

\subsection{VINA correlation}\label{app:vina}
We analyze the correlation between structure-prediction-based objectives and docking-based scores. We observe weak positive correlations between contact loss and Vina score (0.270), predicted affinity and Vina score (0.102), and ipTM and Vina score (0.145). Since contact loss, predicted affinity, and Vina score are all optimized in the lower-is-better direction, the positive correlations for contact loss and affinity indicate weak alignment with docking scores: improving these structure-predictor objectives tends to weakly coincide with better Vina scores, but the relationship is far from deterministic. In contrast, ipTM is a higher-is-better confidence score, whereas Vina is lower-is-better. Therefore, alignment between ipTM and Vina would correspond to a negative correlation. The observed positive correlation instead points in the opposite direction, although its small magnitude indicates that this inverse trend is weak.

Overall, these weak and partly misaligned correlations suggest that structure-prediction-based signals and docking-based scores capture related but distinct aspects of protein--ligand interactions. This limited alignment helps explain why DBMol can achieve strong structure-aligned performance without consistently obtaining the best Vina scores.

\subsection{Similarity to known actives}\label{app:sim}

\begin{table}[ht]
    \centering
    \small
    \caption{\textbf{Similarity to known actives and interaction recovery on LIT-PCBA.}
    Sim measures similarity to the crystallized reference ligand, where lower values indicate less direct copying.
    EF@1\% measures active enrichment in the nearest neighborhood of generated molecules.
    IFP Sim measures residue-level interaction fingerprint similarity to the reference ligand.}
    \setlength{\tabcolsep}{5pt}
    \renewcommand{\arraystretch}{1.05}
    \begin{tabular}{@{}lccc@{}}
    \toprule
    Method & Sim $\downarrow$ & EF@1\% $\uparrow$ & IFP Sim $\uparrow$ \\
    \midrule
    SynFormer & 0.447 & 1.06 & 0.656 \\
    ShEPhERD & 0.137 & 0.80 & 0.401 \\
    SynCoGen & 0.107 & 0.86 & 0.597 \\
    DiffSBDD & 0.106 & 0.86 & 0.622 \\
    CGFlow & 0.093 & 1.46 & 0.600 \\
    DeFoG & 0.101 & 0.81 & 0.343 \\
    \rowcolor{Gray}
    DBMol & 0.094 & 1.33 & 0.392 \\   
    \bottomrule
    \end{tabular}
    \label{tab:active_similarity}
\end{table}

\begin{wraptable}[18]{r}{0.65\textwidth}
    \centering
    \footnotesize
    \caption{\textbf{De novo molecule design results on additional targets with iterative refinement.}
    We compare DBMol with DiffSBDD, a pocket-conditioned structure-based baseline applicable to the strict de novo setting without reference ligands. 
    DBMol (round 2) denotes an exploratory second-round refinement initialized from a first-round DBMol candidate.}
    \setlength{\tabcolsep}{2.8pt}
    \renewcommand{\arraystretch}{1.05}
    \resizebox{1.0\linewidth}{!}{
    \begin{tabular}{@{}lcccccc>{\columncolor{Gray}}c@{}}
    \toprule
    Method 
    & BCov $\uparrow$ 
    & AF3 Succ. $\uparrow$ 
    & iPAE $\downarrow$ 
    & Dist $\downarrow$ 
    & Vina $\downarrow$
    & Div $\uparrow$
    & Avg. Rank $\downarrow$ \\
    \midrule

    \multicolumn{8}{@{}l}{\textit{LGR4}} \\
    DiffSBDD
    & 0.23 & 0.51 & 13.2 & 5.7 & -9.6 & 0.90 & 2.33 \\
    \rowcolor{Gray}
    DBMol
    & 0.25 & 0.71 & 11.4 & 3.9 & -8.1 & 0.87 & \textbf{1.83} \\
    \rowcolor{Gray}
    DBMol (round 2)
    & 0.33 & 0.87 & 11.6 & 3.2 & -7.9 & 0.82 & \textbf{1.83} \\

    \midrule
    \multicolumn{8}{@{}l}{\textit{CD47}} \\
    DiffSBDD
    & 0.01 & 0.06 & 8.3 & 25.3 & -4.0 & 0.93 & 1.83 \\
    \rowcolor{Gray}
    DBMol
    & 0.03 & 0.08 & 8.7 & 28.5 & -6.2 & 0.87 & \textbf{1.67} \\
    \rowcolor{Gray}
    DBMol (round 2)
    & 0.01 & 0.02 & 7.9 & 32.0 & -5.7 & 0.87 & 2.50 \\

    \bottomrule
    \end{tabular}}
    \label{tab:denovo_targets_em2}
    \vspace{-10pt}
\end{wraptable}

We further analyze whether generated molecules simply reproduce known binders on LIT-PCBA. 
EF@1 measures the enrichment of known actives in the top 1\% of ranked candidates over random selection. EF@1\% = 5 means actives are 5 times more enriched in the top 1\% vs random; EF@1\% = 1 is random. IFP is a residue-level interaction fingerprint; IFP similarity between a generated and reference pose quantifies how well the generated pose reproduces the native binding interactions.
As shown in \cref{tab:active_similarity}, DBMol has low similarity to the crystallized reference ligand, with Sim $=0.094$.
This is substantially lower than ligand-guided SynFormer and also lower than the unconditional DeFoG baseline, indicating that DBMol does not directly copy the reference binder.
At the same time, DBMol substantially improves active enrichment over DeFoG, increasing EF@1\% from $0.81$ to $1.33$, and achieves the second-best EF@1\% among the compared methods.
DBMol also modestly improves residue-level interaction recovery over DeFoG, increasing IFP Sim from $0.343$ to $0.392$.
Overall, these results suggest that DBMol tends to generate alternative active-like molecules rather than reproducing known binders, while improving neighborhood-level active enrichment and modestly improving pocket-level interaction similarity. However, the absolute gains are limited, so we report these results only as supplementary analysis rather than including them in the main paper.

\subsection{Iterative refinement from a first-round candidate}\label{app:condition}
We further include an exploratory ablation to test whether DBMol can be reused for iterative refinement.
Starting from an initial DBMol batch, we select the best first-round candidate (the same molecule used to initialize DiffSBDD 3D pocket conditions) according to the Boltz-2-based selection criterion used by DBMol, without using AF3 or Vina evaluation metrics for selection.
We then use this candidate to initialize a second round of optimization and projection.
During this second round, we add a lightweight regularization term that penalizes deviations from the relaxed node and edge distributions of the selected first-round candidate, preventing the optimized molecule from drifting too far from the initialization. Specifically, the loss weight is $0.1$ and penalised the entropy of the current relaxed molecule distribution.
In practice, we regularize the edge distribution more strongly (with doubled loss weight) than the node distribution, which empirically better preserves the molecular scaffold while still allowing local modifications.

As shown in \cref{tab:denovo_targets_em2}, the effect of second-round refinement is target-dependent.
On LGR4, DBMol (round 2) further improves BCov from 0.25 to 0.33, AF3 Success from 0.71 to 0.87, and Dist from 3.9 to 3.2 compared with first-round DBMol, while maintaining a similar average rank.
Its iPAE remains better than DiffSBDD but slightly worse than first-round DBMol, and the main tradeoffs are weaker Vina score and lower molecular diversity.
On CD47, however, the second round does not improve overall performance: first-round DBMol obtains the best average rank, while DBMol (round 2) improves iPAE but degrades BCov, AF3 Success, Dist, and diversity. We further inspected the poor CD47 round-2 result.
The round-2 initialization was selected using only the Boltz-2-based criterion, rather than AF3 metrics, to avoid leakage from the final evaluation.
This selected molecule has 0 AF3 pocket coverage, indicating a poor initialization under the held-out metric.
Since the second round regularizes the relaxed node and edge distributions to stay close to the initialization, the refinement is constrained around an unfavorable molecule, explaining the degraded CD47 performance.

Overall, these results suggest that iterative reuse of DBMol candidates can further improve structural metrics on some targets, but it is not consistently beneficial across targets and is sensitive to the initial molecule.
We therefore treat it as an exploratory refinement strategy rather than as part of the main de novo baseline.
Nevertheless, DBMol remains competitive on de novo targets without relying on reference ligands or requiring a well-defined binding pocket.

\subsection{Runtime and inference-time compute.}\label{app:runtime}
To make the inference-time search effort explicit, we report the runtime of DBMol under the LIT-PCBA protocol in \cref{tab:runtime_litpcba}. 
All LIT-PCBA targets use the same fixed configuration with 50 Boltz-guided optimization steps. 
We report the cost of the DBMol generation pipeline before AF3 evaluation, since AF3 is used only for downstream evaluation and is not part of the molecular optimization objective. 
MSA search is performed once per protein and cached, and is therefore not included in the per-run runtime.

\begin{table}[ht]
    \centering
    \small
    \caption{\textbf{Runtime breakdown of DBMol under the LIT-PCBA protocol.}
    We report the estimated wall-clock time for the fixed 50-step optimization setting used in \cref{tab:main_results}. }
    \setlength{\tabcolsep}{4pt}
    \renewcommand{\arraystretch}{1.05}
    \begin{tabular}{@{}lcc@{}}
    \toprule
    Component & Setting & Runtime \\
    \midrule
    Boltz-guided optimization 
    & 50 steps 
    & $\sim$30 min \\

    DeFoG projection / sampling 
    & per optimized molecule set 
    & $\sim$10 s \\

    \midrule
    DBMol generation total 
    & 50-step protocol 
    & $\sim$30 min \\

    \midrule
    SynFormer projection 
    & optional, 100 molecules 
    & $\sim$25 min \\
    \bottomrule
    \end{tabular}
    \label{tab:runtime_litpcba}
\end{table}

Under this protocol, DBMol uses a fixed number of Boltz calls for every target: one Boltz-guided update per optimization step, for 50 steps in total. 
The dominant cost is therefore the iterative Boltz-guided optimization, which takes approximately 30 minutes under our implementation. 
The subsequent DeFoG projection is comparatively negligible, taking about 10 seconds. 
When the optional DBMol-Syn variant is used, SynFormer projection adds additional runtime, reported separately in \cref{tab:runtime_litpcba}. 
Thus, the additional test-time computation used by DBMol is fixed, target-independent within LIT-PCBA, and explicitly reported.

\begin{table}[ht]
  \centering
  \caption{
    Molecular generation results on the ZINC-250K dataset, comparing DeFoG with other diffusion and flow matching based baselines. Best results are highlighted in bold.
    }
    \resizebox{0.8\linewidth}{!}{
  \label{tab:zinc}
  \begin{tabular}{@{}lccccc@{}}
    \toprule
    \textbf{Model} & Val. $\uparrow$ & Uniqueness $\uparrow$ & FCD {$\downarrow$} & NSPDK {$\downarrow$} & Scaffold {$\uparrow$} \\
    \midrule
    GruM                   & 98.65          & --             & 2.257          & 0.0015        & 0.5299          \\
    GBD                  & 97.87          & --             & 2.248          & 0.0018        & 0.5042          \\
    CatFlow             & 99.21          & \textbf{100.00}         & 13.211         & --            & --              \\
    \midrule
    \rowcolor{Gray}
    DeFoG (w/o charges)                 & {99.22}   & 99.99   & \textbf{1.425}  & \textbf{0.0008} & {0.5903} \\
    \rowcolor{Gray}
    DeFoG (w/ charges)                 & \textbf{99.28}   & \textbf{100.00}   & {1.850}  & {0.0013} & \textbf{0.9281} \\
    \bottomrule
  \end{tabular}}
\end{table}

\subsection{Choice of discrete denoiser}\label{app:defog}
DBMol requires a discrete molecular denoiser that can reliably project optimized relaxed representations back to chemically valid molecules, while remaining computationally efficient.
DeFoG satisfies both requirements.
As shown in Table~\ref{tab:zinc}, DeFoG achieves near-perfect validity and strong distributional alignment on the ZINC-250K benchmark, as measured by chemical validity, uniqueness, FCD, NSPDK, and scaffold diversity.
We further evaluate DeFoG both with and without explicit modeling of formal atomic charges to isolate the effect of charge-aware denoising.
Moreover, DeFoG adopts a flow-matching formulation, enabling high-quality molecular generation with a small number of denoising steps, which is particularly well suited for optimization-in-the-loop pipelines such as DBMol.

\end{document}